\documentclass{article} %
\usepackage{colm2024_conference}

\usepackage{booktabs}
\usepackage{graphicx}
\usepackage{enumitem}
\usepackage{wrapfig}
\usepackage{algorithm}
\usepackage{algpseudocode}

\usepackage{CJKutf8}

\usepackage{microtype}
\usepackage{amsmath}
\usepackage{amsfonts}
\usepackage{colortbl}
\usepackage[utf8]{inputenc}
\usepackage[T1]{fontenc}
\definecolor{lightgray}{rgb}{0.9,0.9,0.9}
\usepackage{caption}
\usepackage{subcaption}
\usepackage{xcolor}
\usepackage{setspace}
\usepackage{url}
\usepackage{multirow}
\usepackage{colortbl}
\usepackage{tabularx}
\usepackage{xltabular}
\usepackage{blindtext}
\usepackage{pgfplots}
\pgfplotsset{compat=1.18} 
\usepackage{tikz}
\usetikzlibrary{er,positioning,bayesnet}
\usepackage{makecell}
\usepackage[safe]{tipa}
\usepackage{siunitx}
\usepackage{nicefrac}
\usepackage{tocloft}
\usepackage{listings}
\usepackage[raster,skins]{tcolorbox} %
\usepackage{xltabular}
\usepackage{adjustbox}
\usepackage{xurl}
\usepackage{multirow}
\usepackage{threeparttable}
\usepackage{xcolor}

\usepackage{amsmath,amsfonts,bm}

\def\eqref#1{equation~\ref{#1}}

\def\1{\bm{1}}

\DeclareMathAlphabet{\mathsfit}{\encodingdefault}{\sfdefault}{m}{sl}
\SetMathAlphabet{\mathsfit}{bold}{\encodingdefault}{\sfdefault}{bx}{n}

\title{Babelfish: Balancing Quality and Latency in SimulST }

\author{
\textbf{Shiyang Deng\textsuperscript{1,2}, Nanjie Li\textsuperscript{1,2}, Haihua Xu\textsuperscript{1}, Wei Shi\textsuperscript{1}, Hao Huang\textsuperscript{2}}\\[0.3em]{\normalfont\small\textsuperscript{1}Timekettle\quad\textsuperscript{2}Xinjiang University}
}

\newcommand{\method}{\textbf{Babelfish}\xspace}

\begin{document}

\maketitle

\begin{abstract}
Simultaneous speech translation requires deciding both when to translate and what to output from an incomplete speech input. We present \method, an open end-to-end model for English$\rightarrow$Chinese and Chinese$\rightarrow$English simultaneous speech-to-text translation, fine-tuned from Qwen3-Omni-30B-A3B-Instruct. Our key idea is to derive streaming supervision from an offline model: Qwen3-Omni scores teacher-generated translations conditioned on progressively longer audio prefixes, and changes in these scores indicate when target content is likely to be supported by the observed speech. We use these signals to construct prefix-to-prefix (P2P) training data without external alignment tools, using pseudo-labels for both translations and emission times. At inference time, \method processes speech in 320\,ms chunks and either emits translation tokens or waits after each chunk, without voice activity detection or a separate read/write policy. A bounded audio window and text context enable translation of unsegmented streams of arbitrary length. Across 11 sentence-level and long-form test sets, \method achieves the highest COMET scores on 10 sets and the highest CometKiwi scores on 9 among the evaluated systems, including Hy-SimulMT and four commercial simultaneous interpretation services. It outperforms Hy-SimulMT in COMET on all 11 sets, with average gains of 2.9 points for English$\rightarrow$Chinese and 2.3 points for Chinese$\rightarrow$English. Average LAAL is 1.74\,s and 2.08\,s for the two directions, respectively. Latency is lower than Hy-SimulMT on 7 sets.
\method also achieves lower reported latency than the
commercial services, although their wall-clock measurements
include network and server-side processing delays.
These results demonstrate a strong translation quality--latency trade-off among the evaluated systems. We release the model weights and inference code.
\end{abstract}

\section{Introduction}

Simultaneous speech translation (SimulST) generates translations while speech is still arriving, enabling cross-lingual communication in meetings, lectures, live broadcasts, and conversations. Unlike offline speech translation, a simultaneous system must repeatedly decide whether the available context is sufficient to produce more output or whether it should wait for additional speech. Early emission reduces latency but risks irreversible errors caused by incomplete context; waiting can improve translation quality but increases delay. Managing this quality--latency trade-off is therefore central to SimulST~\citep{ma2019stacl,ma2020simuleval}.

Prior work has explored several ways to control when a system translates. Explicit read/write policies range from fixed schedules such as wait-$k$~\citep{ma2019stacl} to adaptive mechanisms such as monotonic multihead attention~\citep{ma2020mma}, with extensions from text to speech input~\citep{ma2020simulmt2simulst}. Another line of work adapts offline speech translation models at inference time, using agreement between consecutive hypotheses~\citep{liu2020lowlatency,polak-etal-2022-cuni} or encoder--decoder attention~\citep{papi2023edatt,papi2023alignatt} to determine which translation prefixes to emit. Systems such as SeamlessStreaming~\citep{seamless2023} and StreamSpeech~\citep{zhang2024streamspeech} have further advanced multilingual streaming translation and speech output. Nevertheless, evaluation has largely focused on pre-segmented speech: a survey of 110 SimulST papers found that more than 80\% relied on pre-segmented input~\citep{papi2024real}. Such settings do not fully capture the challenges of translating continuous speech without predefined utterance boundaries.

Large pretrained speech and audio--language models, including Whisper~\citep{radford2022whisper}, Qwen2-Audio~\citep{chu2024qwen2audio}, Qwen2.5-Omni~\citep{xu2025qwen25omni}, and Qwen3-Omni~\citep{xu2025qwen3omni}, provide strong foundations for speech recognition and offline translation. Recent work has increasingly adapted these capabilities to simultaneous translation. CLASI~\citep{cheng2024clasi} and Seed LiveInterpret 2.0~\citep{cheng2025seedliveinterpret} develop end-to-end interpretation systems based on large language models, while InfiniSST~\citep{ouyang2025infinisst} and hierarchical policy optimization~\citep{ouyang2026hpo} address translation of unbounded speech streams. Other approaches use speech-to-text causal alignment for streaming translation and transcription~\citep{koshkin2026streaming} or learn when to commit translations from partial speech~\citep{hoang2026commit}. Hibiki~\citep{labiausse2025hibiki} targets simultaneous translation with high-fidelity speech output, and Hy-SimulMT~\citep{hy_simulmt} provides an openly released model for real-time simultaneous translation. Commercial simultaneous interpretation services have also become available.

Despite these advances, adapting a strong offline model into a practical simultaneous translator presents several challenges. Offline translation supervision specifies what to translate but does not directly specify when each part of the translation can be emitted from partial speech. At inference time, reliance on external segmentation or hand-tuned emission rules can introduce additional tuning requirements and sensitivity to input boundaries. Long-form translation further requires processing continuous speech with bounded context while maintaining continuity in the output. In addition, closed commercial services limit opportunities for reproducible research and self-hosted deployment.

In this report, we present \method, an open end-to-end simultaneous speech-to-text translation model for English$\rightarrow$Chinese and Chinese$\rightarrow$English. Built on Qwen3-Omni-30B-A3B-Instruct~\citep{xu2025qwen3omni}, a mixture-of-experts model with approximately 3B parameters activated per token, \method learns both translation and emission behavior through prefix-to-prefix (P2P) fine-tuning. Our key idea is to derive streaming supervision from the offline model itself. Given a teacher-generated translation, the offline model scores it conditioned on progressively longer audio prefixes. Changes in these scores provide signals for estimating when target content becomes supported by the observed speech. We use these signals to construct P2P training examples, with pseudo-labels for both translation content and emission times, without external alignment tools.

At inference time, \method receives speech in 320\,ms chunks and, after each chunk, either emits additional translation tokens or waits for more input. Emitted text is never revised, and the model operates without voice activity detection, hand-tuned emission thresholds, or a separate read/write policy module. To support unsegmented streams of arbitrary length with bounded context, inference retains the most recent 24\,s of audio and up to 256 tokens of previously emitted translation. An explicit end-of-input signal allows the model to complete any remaining translation.

We evaluate \method on 11 test sets spanning both translation directions and covering sentence-level and long-form speech. These include FLEURS~\citep{conneau2022fleurs}, CoVoST~2~\citep{wang2020covost2}, ACL~60/60~\citep{salesky-etal-2023-evaluating}, the sentence-level and long-form versions of RealSI~\citep{cheng2024clasi}, and the long-form MCIF benchmark~\citep{papi2025mcif}. Compared with Hy-SimulMT and four commercial simultaneous interpretation services, \method achieves the highest COMET~\citep{rei-etal-2022-comet} on 10 test sets and the highest CometKiwi~\citep{rei-etal-2022-cometkiwi} on 9. It outperforms Hy-SimulMT in COMET on all 11 sets, with average gains of 2.9 points for English$\rightarrow$Chinese and 2.3 points for Chinese$\rightarrow$English. Its non-computation-aware LAAL~\citep{papi-etal-2022-generation} ranges from 1.5 to 2.2\,s and is lower than that of Hy-SimulMT on 7 sets. These results demonstrate a strong quality--latency trade-off across the evaluated settings. We release the model weights and inference code to support further research and deployment.

\section{Related Work}
\label{sec:related_work}

\paragraph{Simultaneous translation policies.}
Simultaneous translation requires determining when an incomplete source provides sufficient context to produce additional target text. Fixed schedules such as wait-$k$~\citep{ma2019stacl} and adaptive mechanisms such as monotonic multihead attention~\citep{ma2020mma} explicitly control this read/write decision. An alternative repeatedly applies an offline model to the available input and commits the translation prefix shared by successive hypotheses~\citep{liu2020lowlatency,polak-etal-2022-cuni}. Our work is most closely related to approaches that derive timing supervision from a pretrained model's behavior on partial inputs. These approaches share the goal of estimating whether the observed source supports further translation, but differ in the signals used to construct supervision and the mechanisms used to control emission at inference time.

\paragraph{Divergence and information gain.}
DiG-SST~\citep{chen2024digsst} combines partial-speech training with an adaptive policy supervised by the divergence between translation distributions conditioned on partial and complete input. Its policy module estimates whether additional speech would change the translation distribution sufficiently to warrant waiting. REINA~\citep{hirschkind2026reina} adopts a related information-theoretic formulation: the difference between a reference token's log-probabilities under full and partial audio estimates the information that the remaining speech could provide. This signal, together with regularization, is used to train a separate read/write policy. Both methods derive policy supervision from offline model behavior, while their policy networks operate only on the context available at inference time.

\paragraph{Prefix scores for contextual alignment.}
Hibiki~\citep{labiausse2025hibiki} derives contextual alignments by scoring target words with a text translation model conditioned on progressively longer source-text prefixes. For each target word, it selects the source position associated with the largest increase in conditional log-likelihood. These text-based alignments are mapped to speech through timestamps and used to construct temporally aligned synthetic interpretations, enabling the trained model to generate text and speech with adaptive timing. Our approach shares the use of score changes across source prefixes to estimate when target content becomes supported. The distinction lies in the source of this signal: Hibiki computes alignments in the text domain and transfers them to speech, whereas we score translation tokens directly against growing audio prefixes.

\paragraph{Prefix supervision for speech language models.}
\citet{hoang2026commit} adapt a full-utterance speech language model using translations generated from complete and partial waveforms. Agreement with the full-audio translation determines the target prefix used for supervision, while a synthesis margin allows additional reference tokens to be included based on their probabilities. They compare single-turn forced-prefix decoding with multi-turn append-only decoding and use a confidence threshold to control emission at inference time. Like our approach, this method derives streaming supervision from an offline model's behavior on partial speech. However, it selects supervised prefixes through hypothesis agreement and probability-based extension, whereas we derive emission times from token-score trajectories across growing audio prefixes. Our fine-tuned model also controls emission without an inference-time confidence threshold.

\paragraph{Relation to our approach.}
Our approach builds on the established use of partial-input predictions to supervise translation timing, with a focus on direct audio-based supervision and integrated emission control. Hunyuan generates candidate translations, which are filtered by Qwen3-Omni. A frozen Qwen3-Omni Thinker then scores the retained translation tokens conditioned on growing audio prefixes, and a release strategy converts the resulting score trajectories into prefix-to-prefix training examples for LoRA fine-tuning. The fine-tuned model learns to emit additional translation tokens or wait through its translation and \texttt{EOS} outputs, without a separate policy network or an inference-time confidence threshold. Our contribution is a concrete pipeline that connects audio-conditioned timing supervision with model-controlled emission, together with an evaluation on both sentence-level and unsegmented long-form speech.

\section{Architecture}
\label{sec:arch}

\subsection{Overview}

\begin{figure}[t]
    \centering
    \includegraphics[width=\textwidth]{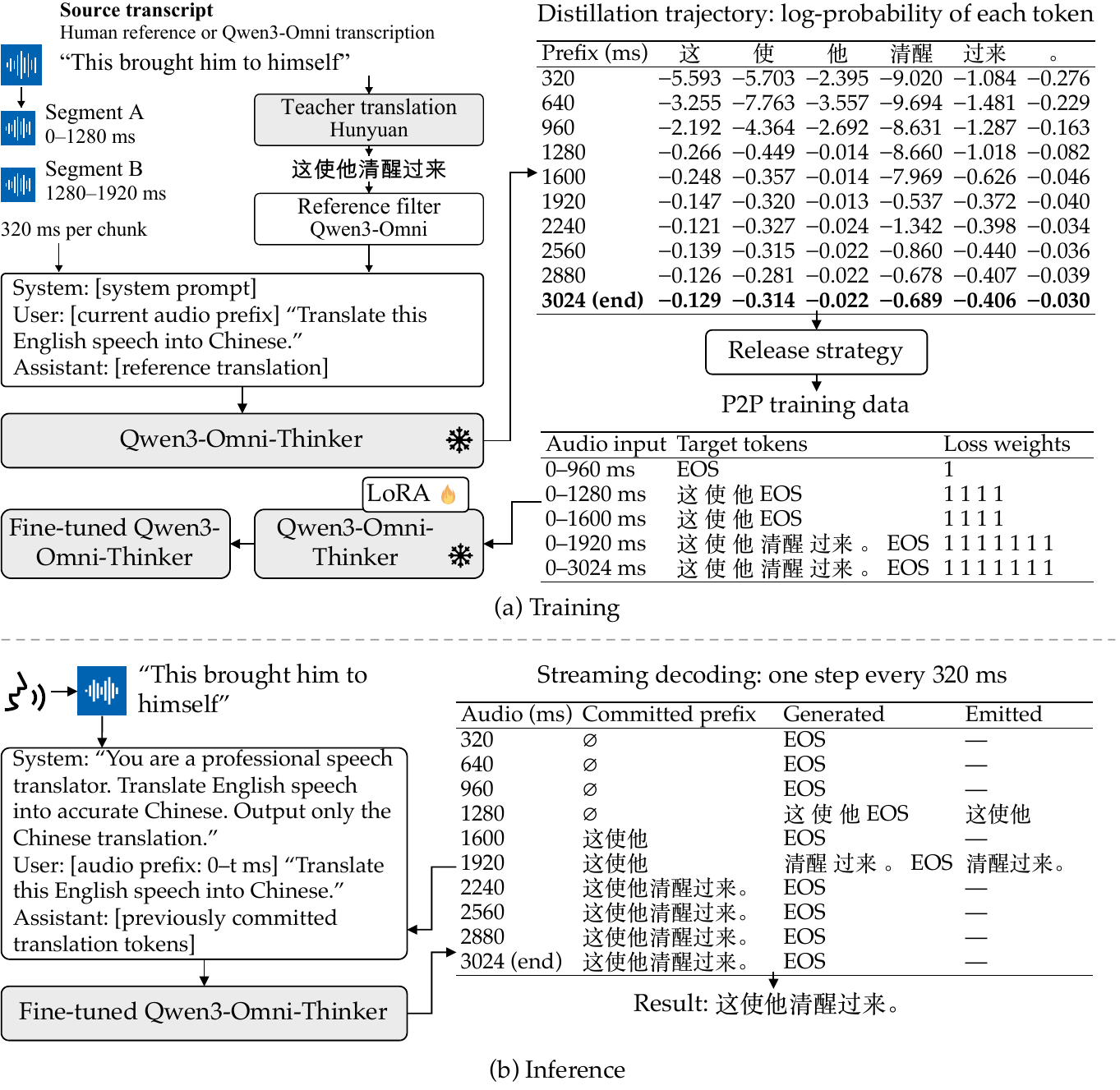}
    \caption{Overview of \method.
    (a) Training: Hunyuan translates source transcripts obtained
    from human annotations or Qwen3-Omni transcription, and
    Qwen3-Omni filters the resulting translations. A frozen
    Qwen3-Omni Thinker scores the retained translation tokens
    conditioned on growing audio prefixes sampled every 320\,ms.
    These score trajectories are converted into emission times
    through threshold crossing, a cumulative maximum in target-token
    order, and alignment to audio-chunk boundaries. The resulting
    prefix-to-prefix (P2P) examples are used for LoRA fine-tuning,
    with unit loss weights for all target tokens, including
    \texttt{EOS}.
    (b) Inference: after each audio chunk, the fine-tuned model
    continues from the previously committed translation.
    An immediate \texttt{EOS} indicates waiting; otherwise,
    newly generated translation tokens are emitted.
    Spaces in the Target tokens and Generated columns indicate
    token boundaries.}
    \label{fig:arch}
\end{figure}

\method adapts an offline audio--language model into a simultaneous
speech-to-text translator through prefix-to-prefix supervision.
As shown in Figure~\ref{fig:arch}, the training pipeline first uses
a frozen offline model to estimate when each translation token
can be released from partial speech. These estimates define
training targets that teach the model both what to translate
and when to stop generating until more audio arrives.

At inference time, a single fine-tuned model performs translation
and emission control. It receives speech in 320\,ms chunks and
continues from the translation already committed. The score-based
release strategy is used only to construct training data:
inference requires neither a separate read/write policy network
nor an emission-confidence threshold.

\subsection{Base Model}

\method is initialized from
Qwen3-Omni-30B-A3B-Instruct~\citep{xu2025qwen3omni}, which follows
a Thinker--Talker architecture. We retain only the Thinker,
consisting of an audio encoder and a mixture-of-experts language
model with approximately 30B parameters, of which approximately
3B are activated per token. The speech-generating Talker is not
used, and the vision encoder is omitted from the released weights.

The model accepts 16\,kHz mono audio and supports
English$\rightarrow$Chinese and Chinese$\rightarrow$English
translation. The translation direction is specified by the
prompt.

\subsection{Streaming Formulation}

Let $\mathbf{x}$ be an input utterance of duration $T$.
We divide the audio into chunks of duration
$\Delta = 320$\,ms, except for a potentially shorter final chunk.
Let $K=\lceil T/\Delta\rceil$ and
$t_k=\min(k\Delta,T)$ denote the audio-prefix boundaries.
We write $\mathbf{x}_{\leq k}$ for the audio observed through
time $t_k$.

At step $k$, the model conditions on the available audio and the
previously committed translation $\hat{\mathbf{y}}^{(k-1)}$.
It generates a continuation terminated by an end-of-turn token,
denoted by \texttt{EOS}. If the first generated token is
\texttt{EOS}, no translation is emitted and the system waits for
more audio. Otherwise, the translation tokens preceding
\texttt{EOS} are appended to the committed output.
Committed tokens are never revised. For an incomplete audio
prefix, \texttt{EOS} ends the current decoding turn rather than
the entire translation stream.

Training and inference use the same chat-style task format.
For English$\rightarrow$Chinese, the system prompt is
\textit{``You are a professional speech translator. Translate
English speech into accurate Chinese. Output only the Chinese
translation.''}
The user turn contains the audio followed by
\textit{``Translate this English speech into Chinese.''}
During inference, the assistant turn is prefilled with the
retained translation context so that generation continues from
the committed output. The Chinese$\rightarrow$English direction
uses the corresponding prompts with the languages exchanged.

\subsection{Training Data Construction}

The training pipeline consists of reference generation and
filtering, token scoring on partial audio, and conversion of the
resulting trajectories into P2P examples.

\paragraph{Reference translations.}
For each training utterance, the source transcript is obtained
from a human annotation or generated by Qwen3-Omni.
Hunyuan translates this transcript into the target language,
and Qwen3-Omni filters the resulting translations.
The retained teacher-generated translations serve as reference
targets for the subsequent scoring and fine-tuning stages.

\paragraph{Prefix scoring.}
Let $\mathbf{y}=(y_1,\ldots,y_m)$ be a retained reference
translation. A frozen offline Qwen3-Omni Thinker scores every
reference token under each audio prefix using teacher forcing:
\begin{equation}
    s_{k,j}
    =
    \log p_{\theta_0}
    \left(
        y_j \mid \mathbf{x}_{\leq k}, y_{<j}
    \right),
\end{equation}
where $\theta_0$ denotes the frozen model parameters and the
fixed task prompt is omitted for brevity.
The reference history $y_{<j}$ is supplied when scoring token
$y_j$; the model does not freely generate a new translation for
each prefix.

The sequence $(s_{1,j},\ldots,s_{K,j})$ describes how the score
of token $y_j$ changes as more speech becomes available.
We refer to these sequences collectively as the
\emph{distillation trajectory}.
For example, in Figure~\ref{fig:arch}a, the score of
\begin{CJK}{UTF8}{gbsn}清醒\end{CJK}
increases from $-7.969$ at 1600\,ms to $-0.537$ at 1920\,ms.
Such changes provide a signal for estimating emission times.
However, scores need not increase monotonically, and some tokens
may already be predictable from the reference text history
before the relevant speech is available.

\paragraph{Emission-time estimation.}
We first estimate an emission time independently for each
target token by comparing its prefix-conditioned score with
its full-audio score $s_{K,j}$.
Let $\bar{s}_j(t)$ denote the piecewise-linear interpolation
of the scores at the sampled audio-prefix boundaries.
The initial time estimate is the first threshold crossing:
\begin{equation}
    \tilde{\tau}_j
    =
    \inf
    \left\{
        t \in [t_1,T] :
        \bar{s}_j(t) + \delta \geq s_{K,j}
    \right\},
    \qquad \delta = 1.0.
\end{equation}
Thus, a token becomes eligible when its score is no more than
$\delta$ below its full-audio score. If it already satisfies
the condition at the first sampled prefix, its initial time
is $t_1$.

These independently estimated times may violate target-token
order. We therefore apply a cumulative maximum from left to
right:
\begin{equation}
    \tau_j
    =
    \max_{1 \leq i \leq j}\tilde{\tau}_i.
\end{equation}
This enforces $\tau_j \geq \tau_{j-1}$, ensuring that no token
is released before a preceding token. The monotonicity
constraint applies to emission times, not to the underlying
score trajectories.

Finally, each time is rounded up to the next available
audio-prefix boundary:
\begin{equation}
    \hat{\tau}_j
    =
    \min
    \left\{
        t_k : t_k \geq \tau_j
    \right\}.
\end{equation}
In the example in Figure~\ref{fig:arch}, this procedure assigns
the first three tokens to 1280\,ms and the remaining three
tokens to 1920\,ms. Although
\begin{CJK}{UTF8}{gbsn}过来\end{CJK}
and the final punctuation satisfy the threshold earlier, the
cumulative maximum delays their release until
\begin{CJK}{UTF8}{gbsn}清醒\end{CJK}
can also be emitted.

\paragraph{Prefix-to-prefix examples.}
The number of target tokens released at audio-prefix boundary
$t_k$ is
\begin{equation}
    r_k
    =
    \sum_{j=1}^{m}
    \mathbb{I}\mkern-3mu\left[\hat{\tau}_j \leq t_k\right].
\end{equation}
Because emission times are non-decreasing in target-token
order, the released tokens always form a contiguous target
prefix. Moreover, $r_k$ is non-decreasing over audio steps,
and $r_K=m$.

Each audio prefix is paired with the target sequence
\begin{equation}
    \mathbf{z}^{(k)}
    =
    \left(y_1,\ldots,y_{r_k},\texttt{EOS}\right).
\end{equation}
When $r_k=0$, the target consists only of \texttt{EOS}.
When additional audio does not release any new tokens,
successive examples retain the same target prefix.
For instance, the 1280\,ms and 1600\,ms prefixes in
Figure~\ref{fig:arch} share the target
\begin{CJK}{UTF8}{gbsn}这使他\end{CJK}
followed by \texttt{EOS}.

We use standard token-level cross-entropy:
\begin{equation}
    \mathcal{L}_k
    =
    -\sum_{j=1}^{r_k+1}
    \log p_{\theta}
    \left(
        z_j^{(k)}
        \mid
        \mathbf{x}_{\leq k},
        z_{<j}^{(k)}
    \right).
\end{equation}
All target tokens receive unit weight, including \texttt{EOS}
for both incomplete and complete audio.
The \texttt{EOS} targets teach the model where to end each
decoding turn given the currently available speech.

\subsection{Fine-tuning}

We fine-tune the model on the P2P examples using
LoRA~\citep{hu2021lora}. The pretrained parameters remain
frozen, and only the low-rank adapters are optimized.
After training, the adapters are merged into the retained
Thinker weights. The resulting model preserves the Thinker
backbone and requires no additional policy network for
streaming translation.

\subsection{Streaming Inference}

Speech arrives in 320\,ms chunks.
After each chunk, the model receives the most recent 24\,s
of audio. Its assistant turn is prefilled with the committed
translation associated with that audio, preceded by up to
256 earlier translation tokens as text context.
The retained text context begins at a sentence boundary.

The model generates until \texttt{EOS} or a limit of 96 new
tokens is reached. An immediate \texttt{EOS} produces no
new output; otherwise, the generated translation tokens
are committed after the repetition check described below.
At the next step, decoding continues from the updated
translation prefix. Neither the offline scoring model nor
the training-time release strategy is used during inference.

The sliding audio window and limited preceding text history
allow the system to process unsegmented streams without
retaining the full input history.
After the final audio chunk, we make an additional decoding
call with the marker \texttt{[audio complete]} appended to
the user instruction, prompting the model to finish any
remaining translation rather than wait for further speech.

We use greedy decoding by default and also support beam
search with two beams. A lightweight repetition guard checks
each newly generated continuation before it is committed.
If runaway repetition is detected, the continuation is
truncated at the start of the repetition, for example before
a phrase would be repeated for the third time.
Continuations without detected repetition are unchanged.

\section{Evaluation}
\label{sec:eval}

\paragraph{Test sets.}
We evaluate \method on 11 test sets across
English$\rightarrow$Chinese and Chinese$\rightarrow$English,
covering both sentence-level and long-form speech
(Table~\ref{tab:testsets}).
The long-form benchmarks comprise MCIF~\citep{papi2025mcif}
and RealSI-L~\citep{cheng2024clasi}.
The sentence-level benchmarks comprise RealSI-S, obtained by
splitting RealSI recordings at reference segment boundaries;
FLEURS~\citep{conneau2022fleurs};
CoVoST~2~\citep{wang2020covost2}; and the development and
evaluation splits of ACL~60/60~\citep{salesky-etal-2023-evaluating}.
Each translation direction and each ACL~60/60 split is counted
as a separate test set.

\begin{table}[t]
    \centering
    \caption{Evaluation datasets and numbers of input examples
    per translation direction. Counts refer to full recordings
    for long-form datasets and individual segments for
    sentence-level datasets. A dash indicates a direction
    not evaluated.}
    \label{tab:testsets}
    \begin{tabular}{@{}llrr@{}}
        \toprule
        \textbf{Test set} &
        \textbf{Speech} &
        \textbf{En$\rightarrow$Zh} &
        \textbf{Zh$\rightarrow$En} \\
        \midrule
        \multicolumn{4}{@{}l}{\textit{Long-form}} \\
        MCIF
            & Scientific talks
            & 21 & -- \\
        RealSI-L
            & Spontaneous speech, 10 domains
            & 10 & 10 \\
        \midrule
        \multicolumn{4}{@{}l}{\textit{Sentence-level}} \\
        RealSI-S
            & RealSI split at reference boundaries
            & 346 & 431 \\
        FLEURS
            & Read speech
            & 646 & 945 \\
        CoVoST~2
            & Read speech
            & 15,530 & 4,898 \\
        ACL~60/60 dev / eval
            & Talks from ACL 2022
            & 468 / 416 & -- \\
        \bottomrule
    \end{tabular}
\end{table}

\paragraph{Quality metrics.}
We measure translation quality using COMET
(\texttt{wmt22-comet-da})~\citep{rei-etal-2022-comet},
reference-free CometKiwi
(\texttt{wmt22-cometkiwi-da})~\citep{rei-etal-2022-cometkiwi},
and BLEU~\citep{papineni-etal-2002-bleu}.
BLEU is computed with SacreBLEU~\citep{post-2018-call},
using the \texttt{zh} tokenizer for Chinese output and
\texttt{13a} for English output.
COMET and CometKiwi scores are multiplied by 100.
Outputs are evaluated as produced, including punctuation,
and empty outputs are retained as empty strings for
quality scoring.

Sentence-level outputs are evaluated against their
corresponding references.
For long-form recordings, we first re-segment each output
to match the reference sentences using
OmniSTEval~\citep{polak2025betterlate}, then compute metrics
on the resulting segments.

\paragraph{Latency metrics.}
For \method and Hy-SimulMT, we report non-computation-aware
LAAL~\citep{papi-etal-2022-generation}.
The emission timestamp of each output unit is determined by
the duration of audio consumed when that unit is emitted,
excluding model computation time.
We use characters as output units for Chinese and words
for English; an English word is considered emitted when
its final character is emitted.
For long-form recordings, latency is evaluated after
re-segmentation using StreamLAAL~\citep{papi2024streamatt}.
Empty outputs are excluded from latency calculations.

For commercial services, emission timestamps are measured
from the wall-clock arrival times of streamed text.
Their latency therefore includes network and server-side
processing delays, whereas the offline streaming simulations
for \method and Hy-SimulMT exclude these factors.
The two latency settings are not directly comparable.

\paragraph{Baselines.}
We compare \method with the openly released
Hy-SimulMT-30B-A3B~\citep{hy_simulmt} and four commercial
simultaneous interpretation services.
Hy-SimulMT is evaluated through offline streaming simulation
using the default configuration of its official inference
code, with 1.6\,s chunks and voice activity detection.
Its latency is computed using the same audio-consumption
timestamps as \method.

The commercial systems are Qwen3.8-LiveTranslate
(\texttt{qwen3.8-livetranslate-flash-realtime}),
Seed LiveInterpret 2.0~\citep{cheng2025seedliveinterpret},
GPT-Realtime-Translate
(\texttt{gpt-realtime-translate}, version 2026-05-06
on Microsoft Foundry), and Gemini 3.5 Live Translate
(\texttt{gemini-3.5-live-translate-preview}).
All services were evaluated in speech-to-text mode through
their streaming APIs between 30 October and 1 November 2026,
with audio transmitted in real time.
Requests originated from mainland China; GPT-Realtime-Translate
and Gemini were accessed through proxies.

\paragraph{Incomplete results and latency adjustment.}
GPT-Realtime-Translate returned valid results for only a subset
of CoVoST~2.
For this dataset, we estimate its full-set quality scores by
adding to its observed subset scores the mean full-set-minus-subset
score difference across the other five systems.
These estimates are marked with $^*$; the COMET adjustments
are approximately within $\pm0.5$ points.
They are extrapolated estimates rather than measurements
on the complete test set.

For GPT-Realtime-Translate on CoVoST~2
English$\rightarrow$Chinese, we subtract approximately
0.54\,s from the measured latency to account for a delay
attributed to server overload.
Its Chinese outputs contain English punctuation, which is
retained under the same scoring protocol used for all systems.

\paragraph{Decoding.}
We evaluate \method through offline streaming simulation
using vLLM~\citep{kwon2023vllm} in FP16 with beam width 2.
Decoding follows Section~\ref{sec:arch}, including 320\,ms
input steps, a 24\,s audio window, up to 256 earlier
translation tokens as text context, and the final
end-of-input marker.

The repetition guard is disabled for the reported results;
enabling it changes only one of the 23,721 outputs.
We decode multiple recordings in batches.
Small numerical differences associated with batch composition
lead to run-to-run variation of approximately
$\pm0.1$ COMET points.

\subsection{Results}

\begin{table}[t]
\caption{\textbf{Results on the 11 test sets.} Each block reports one metric; $\uparrow$: higher is better, $\downarrow$: lower is better. COMET and CometKiwi are scaled to 0--100. LAAL is in seconds and is computed as StreamLAAL on the long-form sets (MCIF and RealSI L). L and S denote the long-form and sentence-level versions of RealSI; ACL denotes ACL 60/60. The best value in each column of each block is in bold. $^\dagger$Commercial service; its latency is wall-clock time measured in real time, including network round trips and server-side computation, and is not directly comparable with the offline-simulated latency of the first two systems. $^*$Approximate (see text).}\label{tab:main}\small\setlength{\tabcolsep}{3pt}\begin{adjustbox}{max width=\textwidth}\begin{tabular}{@{}lccccccc@{\hspace{9pt}}cccc@{}}\toprule & \multicolumn{7}{c}{\textbf{English$\rightarrow$Chinese}} & \multicolumn{4}{c}{\textbf{Chinese$\rightarrow$English}} \\\cmidrule(lr){2-8} \cmidrule(l){9-12}\multirow{2}{*}{\textbf{System}} & \multirow{2}{*}{MCIF} & \multicolumn{2}{c}{RealSI} & \multirow{2}{*}{FLEURS} & \multirow{2}{*}{CoVoST~2} & \multicolumn{2}{c}{ACL} & \multicolumn{2}{c}{RealSI} & \multirow{2}{*}{FLEURS} & \multirow{2}{*}{CoVoST~2} \\\cmidrule(lr){3-4} \cmidrule(lr){7-8} \cmidrule(lr){9-10} & & L & S & & & dev & eval & L & S & & \\\midrule\midrule\multicolumn{12}{@{}l}{\textit{COMET $\uparrow$}} \\\midrule\method & \textbf{86.8} & \textbf{82.8} & \textbf{82.9} & \textbf{88.3} & \textbf{87.1} & \textbf{85.7} & \textbf{85.2} & 77.8 & \textbf{79.5} & \textbf{85.7} & \textbf{80.6} \\Hy-SimulMT-30B-A3B & 85.2 & 79.4 & 77.7 & 86.7 & 83.9 & 83.3 & 82.5 & 75.1 & 75.9 & 85.2 & 78.3 \\Qwen3.8-LiveTranslate$^\dagger$ & 86.6 & 82.2 & 80.4 & 86.9 & 85.6 & 84.3 & 84.0 & 77.5 & 77.7 & 84.4 & 79.0 \\Seed LiveInterpret 2.0$^\dagger$ & 86.2 & 81.7 & 79.2 & 85.8 & 81.2 & 83.3 & 81.2 & \textbf{78.0} & 77.8 & 84.5 & 75.2 \\GPT-Realtime-Translate$^\dagger$ & 84.1 & 77.5 & 77.6 & 83.0 & 75.0$^*$ & 80.6 & 80.6 & 74.5 & 73.3 & 79.6 & 65.8$^*$ \\Gemini 3.5 Live Translate$^\dagger$ & 83.1 & 76.3 & 75.6 & 84.0 & 80.4 & 80.8 & 78.9 & 72.2 & 72.3 & 80.4 & 66.0 \\\midrule\midrule\multicolumn{12}{@{}l}{\textit{CometKiwi $\uparrow$}} \\\midrule\method & \textbf{79.7} & \textbf{73.2} & \textbf{75.5} & \textbf{84.2} & \textbf{80.8} & \textbf{78.4} & \textbf{76.6} & 67.8 & 73.5 & \textbf{83.6} & \textbf{77.3} \\Hy-SimulMT-30B-A3B & 76.9 & 70.4 & 70.6 & 81.9 & 76.8 & 74.4 & 72.5 & \textbf{68.7} & 72.5 & 82.5 & 74.1 \\Qwen3.8-LiveTranslate$^\dagger$ & 78.5 & 71.3 & 72.1 & 81.3 & 78.0 & 75.3 & 73.0 & 67.9 & \textbf{73.9} & 81.9 & 74.8 \\Seed LiveInterpret 2.0$^\dagger$ & 77.5 & 71.4 & 69.7 & 80.4 & 72.4 & 73.7 & 69.7 & 67.0 & 69.5 & 81.4 & 69.9 \\GPT-Realtime-Translate$^\dagger$ & 76.7 & 67.5 & 69.8 & 78.0 & 66.7$^*$ & 72.5 & 70.9 & 65.1 & 67.8 & 76.2 & 59.8$^*$ \\Gemini 3.5 Live Translate$^\dagger$ & 74.2 & 67.8 & 67.7 & 78.3 & 72.0 & 70.7 & 67.8 & 65.6 & 67.2 & 75.8 & 59.7 \\\midrule\midrule\multicolumn{12}{@{}l}{\textit{BLEU $\uparrow$}} \\\midrule\method & 40.2 & 25.2 & 24.0 & 40.3 & \textbf{41.4} & 41.8 & 41.9 & 21.0 & 20.3 & 22.7 & \textbf{23.6} \\Hy-SimulMT-30B-A3B & 52.3 & 29.8 & 28.5 & 42.6 & 38.2 & 49.9 & 48.3 & 22.1 & 19.7 & \textbf{23.7} & 22.9 \\Qwen3.8-LiveTranslate$^\dagger$ & 45.1 & 27.0 & 25.3 & 38.2 & 38.1 & 43.1 & 43.0 & 20.4 & 19.5 & 20.7 & 21.5 \\Seed LiveInterpret 2.0$^\dagger$ & \textbf{55.6} & \textbf{33.3} & \textbf{31.3} & \textbf{43.1} & 39.6 & \textbf{51.9} & \textbf{48.9} & \textbf{28.3} & \textbf{25.3} & 22.7 & 22.2 \\GPT-Realtime-Translate$^\dagger$ & 38.4 & 25.0 & 22.8 & 31.1 & 28.0$^*$ & 33.8 & 34.6 & 17.2 & 14.4 & 16.1 & 11.8$^*$ \\Gemini 3.5 Live Translate$^\dagger$ & 48.3 & 28.4 & 27.0 & 40.3 & 38.7 & 46.5 & 46.1 & 16.8 & 16.8 & 19.4 & 14.6 \\\midrule\midrule\multicolumn{12}{@{}l}{\textit{LAAL (s) $\downarrow$}} \\\midrule\method & \textbf{1.50} & \textbf{1.74} & \textbf{1.75} & \textbf{1.66} & \textbf{1.81} & 1.81 & 1.92 & 1.97 & 2.15 & \textbf{2.07} & \textbf{2.11} \\Hy-SimulMT-30B-A3B & 2.15 & 2.66 & 1.88 & 2.09 & 1.96 & \textbf{1.77} & \textbf{1.74} & \textbf{1.91} & \textbf{1.59} & 2.30 & 2.20 \\Qwen3.8-LiveTranslate$^\dagger$ & 4.07 & 3.69 & 2.18 & 2.30 & 2.66 & 2.44 & 2.51 & 3.56 & 2.42 & 2.73 & 3.24 \\Seed LiveInterpret 2.0$^\dagger$ & 2.51 & 2.78 & 2.02 & 2.15 & 1.86 & 2.10 & 2.02 & 2.82 & 2.19 & 2.70 & 2.13 \\GPT-Realtime-Translate$^\dagger$ & 4.50 & 4.86 & 2.92 & 3.55 & 3.31$^*$ & 2.87 & 2.92 & 4.15 & 3.10 & 3.56 & 3.79$^*$ \\Gemini 3.5 Live Translate$^\dagger$ & 1.98 & 2.24 & 2.74 & 2.53 & 2.95 & 2.67 & 2.55 & 2.06 & 2.80 & 2.82 & 3.05 \\\bottomrule\end{tabular}\end{adjustbox}\end{table}


\paragraph{Overall.}
Table~\ref{tab:main} summarizes the results on all 11 test sets.
\method achieves the highest COMET on 10 sets and the highest
CometKiwi on 9.
Its average COMET across test sets is 85.5 for
English$\rightarrow$Chinese and 80.9 for
Chinese$\rightarrow$English, compared with 84.3 and 79.7
for Qwen3.8-LiveTranslate, the strongest baseline by average
COMET, and 82.7 and 78.6 for Hy-SimulMT.
Under the same non-computation-aware latency protocol,
\method also achieves lower latency than Hy-SimulMT on
7 sets: MCIF, RealSI-L, RealSI-S, FLEURS, and CoVoST~2
for English$\rightarrow$Chinese, and FLEURS and CoVoST~2
for Chinese$\rightarrow$English.
On each of these sets, \method combines the highest COMET
among all evaluated systems with lower latency than
the open baseline.

\paragraph{Comparison with Hy-SimulMT.}
Hy-SimulMT is the only other open model in our comparison.
Both models are evaluated under the same offline streaming
protocol, allowing direct comparison of translation quality
and non-computation-aware latency.

\method outperforms Hy-SimulMT in COMET on all 11 test sets,
with gains ranging from 0.5 points on
Chinese$\rightarrow$English FLEURS to 5.2 points on
English$\rightarrow$Chinese RealSI-S.
The average gains are 2.9 points for
English$\rightarrow$Chinese and 2.3 points for
Chinese$\rightarrow$English.
Improvements are particularly pronounced on RealSI:
3.4 and 5.2 points on the English$\rightarrow$Chinese
long-form and sentence-level sets, respectively, and
2.7 and 3.6 points on their Chinese$\rightarrow$English
counterparts.
CometKiwi also favours \method on 10 sets, with gains
of 1.0--4.9 points.
The exception is Chinese$\rightarrow$English RealSI-L,
where Hy-SimulMT scores 0.9 points higher.

These quality gains are accompanied by lower latency on
7 of the 11 sets, with reductions of 0.09--0.92\,s.
The largest reductions occur on the long-form
English$\rightarrow$Chinese benchmarks:
latency decreases from 2.15 to 1.50\,s on MCIF and
from 2.66 to 1.74\,s on RealSI-L.
On the remaining four sets---the two ACL~60/60 splits
and both Chinese$\rightarrow$English RealSI sets---\method
has 0.04--0.56\,s higher latency, while improving COMET
by 2.4--3.6 points.
Thus, \method improves COMET throughout the comparison
and achieves both higher COMET and lower latency on
a majority of the test sets.

\paragraph{Comparison with commercial services.}
Qwen3.8-LiveTranslate is the strongest commercial baseline
in terms of overall COMET performance.
It ranks second among all six systems on 8 of the 11 sets,
including every English$\rightarrow$Chinese set and
Chinese$\rightarrow$English CoVoST~2.
It comes closest to \method on MCIF and
English$\rightarrow$Chinese RealSI-L, trailing by
0.2 and 0.6 points, respectively.
On the remaining sets, \method leads it by
1.2--2.5 points for English$\rightarrow$Chinese and
0.3--1.8 points for Chinese$\rightarrow$English.

Seed LiveInterpret 2.0 is competitive on long-form speech.
It trails \method by 0.6 points on MCIF and 1.1 points on
English$\rightarrow$Chinese RealSI-L, and leads by
0.2 points on Chinese$\rightarrow$English RealSI-L.
Its performance is less competitive on CoVoST~2,
where \method leads by 5.9 points for
English$\rightarrow$Chinese and 5.4 points for
Chinese$\rightarrow$English.
GPT-Realtime-Translate and Gemini 3.5 Live Translate
trail \method by 2.7 to approximately 14.8 COMET points,
with the largest gaps on Chinese$\rightarrow$English
CoVoST~2.
The GPT-Realtime-Translate results on CoVoST~2 include
the extrapolated estimates described in the evaluation
setup.

The reported commercial-service latencies range from
1.86 to 4.86\,s and are numerically higher than those
of \method on every test set.
However, they include network and server-side processing
delays, whereas the latency of \method excludes computation
time.
These measurements describe the services under the tested
access conditions and do not establish an intrinsic latency
advantage for \method.

\paragraph{Reference-free evaluation.}
CometKiwi provides a complementary assessment that does
not use a reference translation.
Its results broadly support the COMET findings:
\method ranks first on 9 of the 11 test sets.
It leads the next-best system by 1.2--3.6 points on all
English$\rightarrow$Chinese sets, and by 1.1 and 2.5 points
on Chinese$\rightarrow$English FLEURS and CoVoST~2,
respectively.

The two exceptions are the Chinese$\rightarrow$English
RealSI sets.
On RealSI-L, \method ranks third, behind Hy-SimulMT
by 0.9 points and Qwen3.8-LiveTranslate by 0.1 points.
On RealSI-S, it trails Qwen3.8-LiveTranslate by 0.4 points.
The agreement between reference-based and reference-free
evaluation strengthens the evidence for the gains on most
benchmarks, while both metrics identify
Chinese$\rightarrow$English RealSI-L as a weaker setting.

\paragraph{BLEU.}
BLEU produces a different ranking.
Seed LiveInterpret 2.0 ranks first on 8 sets and Hy-SimulMT
on one, whereas \method ranks first only on CoVoST~2
in both directions.
The average rank of \method among the six systems is
3.4 under BLEU, compared with 1.1 under COMET and
1.3 under CometKiwi.

The disagreement is most pronounced on the
English$\rightarrow$Chinese talks in MCIF and ACL~60/60.
On these sets, \method ranks first under both neural
metrics but fifth under BLEU, trailing
Seed LiveInterpret 2.0 by 7.0--15.4 points.
Similarly, Hy-SimulMT has higher BLEU than \method
on 8 sets despite lower COMET on all 11.

BLEU measures reference n-gram overlap and is sensitive
to lexical choices and phrasing.
With Chinese tokenization, this comparison is largely
based on character n-grams.
Previous evaluations have found stronger agreement with
human judgements for neural metrics such as
COMET~\citep{freitag-etal-2022-results}.
The results are therefore consistent with differences
in wording contributing to the lower BLEU scores.
However, agreement between COMET and CometKiwi does not
rule out omissions or other translation errors.
Human evaluation is needed to determine the extent to
which the metric disagreement reflects acceptable
paraphrasing rather than differences in translation quality.

\paragraph{Latency.}
The non-computation-aware latency of \method ranges from
1.50 to 1.92\,s for English$\rightarrow$Chinese and
from 1.97 to 2.15\,s for Chinese$\rightarrow$English,
with averages of 1.74 and 2.08\,s, respectively.
Sentence-level results use LAAL, while long-form results
use StreamLAAL following re-segmentation.

We observe no systematic latency increase on the
long-form benchmarks.
Latency is 1.50\,s on MCIF, 1.74\,s on
English$\rightarrow$Chinese RealSI-L, and 1.97\,s on
Chinese$\rightarrow$English RealSI-L.
These values are below or within the corresponding
sentence-level ranges of 1.66--1.92\,s and 2.07--2.15\,s.
This supports the model's ability to maintain low
algorithmic latency on the evaluated continuous streams.
It does not, by itself, establish that latency is
independent of recording duration, since the benchmarks
also differ in content and speaking style.
The bounded-context inference design limits the history
processed at each step, but does not directly guarantee
a particular emission delay.

\paragraph{Long-form speech.}
On the three long-form sets, \method achieves the highest
COMET on MCIF (86.8) and
English$\rightarrow$Chinese RealSI-L (82.8).
On Chinese$\rightarrow$English RealSI-L, it scores 77.8,
slightly below Seed LiveInterpret 2.0 at 78.0.

RealSI-L and RealSI-S contain the same speech presented
as continuous recordings and pre-segmented sentences,
respectively.
Their comparison therefore provides a useful view of
sensitivity to input segmentation, although the long-form
scores also depend on output re-segmentation.
For English$\rightarrow$Chinese, \method performs almost
identically in the two settings:
82.8 versus 82.9 COMET, with latency of
1.74 versus 1.75\,s.

For Chinese$\rightarrow$English, however, COMET decreases
from 79.5 on RealSI-S to 77.8 on RealSI-L.
This 1.7-point decrease is larger than that of any other
evaluated system, whose decreases are at most 0.8 points.
Chinese$\rightarrow$English RealSI-L is also the only
test set on which \method does not achieve the highest
COMET.
Together, these results identify long-form
Chinese$\rightarrow$English translation as the clearest
priority for further improvement.

\section{Conclusion}
\label{sec:conclusion}

We presented \method, an open end-to-end model for simultaneous
speech-to-text translation between English and Chinese.
Our approach derives timing supervision from a frozen
Qwen3-Omni model by scoring translation tokens on growing
audio prefixes. A threshold-based release strategy with
monotonicity constraints converts these score trajectories
into prefix-to-prefix training examples, without external
alignment tools. LoRA fine-tuning enables the model to learn
both translation and emission behavior. At inference time,
the model processes speech in 320\,ms steps and generates
committed translations from continuous, unsegmented input,
using a sliding audio window and limited text history.
No separate read/write policy or emission-confidence
threshold is required.

Across 11 sentence-level and long-form test sets, \method
achieves the highest COMET on 10 sets and the highest
CometKiwi on 9 among the evaluated systems.
It outperforms Hy-SimulMT in COMET on all 11 sets while
achieving lower non-computation-aware latency on 7.
Average latency is 1.74\,s for English$\rightarrow$Chinese
and 2.08\,s for Chinese$\rightarrow$English.
On English$\rightarrow$Chinese RealSI, translation quality
and latency remain nearly unchanged when moving from
pre-segmented sentences to complete recordings.
These findings support prefix-score supervision as a
practical means of adapting an offline audio--language
model to simultaneous translation.

Future work will focus on improving long-form
Chinese$\rightarrow$English translation and examining
how release strategies and training objectives affect
the quality--latency trade-off.
Human evaluation is needed to resolve the differences
between neural-metric and BLEU rankings.
End-to-end latency comparisons with commercial services
also require computation-aware measurements under matched
deployment conditions.
We release the model weights and inference code to support
reproducible research and further development of open
simultaneous translation systems.

\par\noindent\begin{minipage}{\linewidth}
\section{Acknowledgements}
\label{sec:acknowledgements}

We would like to thank the entire team for their collective efforts and contributions, especially Jun Guo and Yuanming Luo for their assistance with data preparation, and Xiaoyong Guo and Xiansu Deng for their valuable discussions and support with model training.
\end{minipage}\par

\clearpage

\clearpage
\bibliography{biblio}
\bibliographystyle{colm2024_conference}

\end{document}